\documentclass{acm_proc_article-sp} 

\begin{document}
%
\conferenceinfo{Digital Futures `12}{October 23 - 25, 2012, Aberdeen, UK}
\CopyrightYear{2012} 
\crdata{}  

\title{Finding the creatures of habit; Clustering households based on their flexibility in using electricity}

%
%
%
%
%

\numberofauthors{2} 
%
\author{
%
%
\alignauthor
Ian Dent, Uwe Aickelin and Tom Rodden\titlenote{Ian Dent is the corresponding author.}\\
       \affaddr{School of Computer Science}\\
       \affaddr{University of Nottingham}\\
       \affaddr{Nottingham, UK, NG8 1BB}\\
       \email{psxid@nottingham.ac.uk}      
\alignauthor
Tony Craig\\
       \affaddr{The James Hutton Institute}\\
       \affaddr{Craigiebuckler}\\
       \affaddr{Aberdeen, UK, AB15 8QH}\\
       \email{tony.craig@hutton.ac.uk}  
} 
\date{15th November 2011}

\maketitle
\begin{abstract}
Changes in the UK electricity market, particularly with the roll out of smart meters, will provide greatly increased opportunities for initiatives intended to change households' electricity usage patterns for the benefit of the overall system. Users show differences in their regular behaviours and clustering households into similar groupings based on this variability provides for efficient targeting of initiatives. Those people who are stuck into a regular pattern of activity may be the least receptive to an initiative to change behaviour. A sample of 180 households from the UK are clustered into four groups as an initial test of the concept and useful, actionable groupings are found.
\end{abstract}




\section{Introduction}
The electricity market in the UK is undergoing dramatic changes. Legal, social and political drives for a more carbon efficient electricity network, along with the dramatically increased flow of data from households through the deployment of smart meters, is leading to a disruptive change in existing practices. In particular, the change of sampling of electricity usage from a 3 monthly billing cycle to a 30 minute (or shorter) sampling period using smart meters, alters the degree of understanding of households' behaviour that is possible \cite{Energy2009}.

One approach to addressing the pressures on the electricity network is the application of Demand Side Management (DSM) techniques to achieve changes in consumer behaviour at a domestic level \cite{river2005primer}. The Desimax project \cite{Kiprakis2011} (of which this work forms part) concentrates on applying DSM techniques to achieve better electricity system efficiency. The peak time for electricity usage in the UK is during the early evening and the successful application of techniques to reduce, or move, the peak usage would improve the efficiency of the electricity network. 

A marketing practice to efficiently address the most receptive customers is to partition the customers into a few (generally fewer than 10) stereotypical customers which can then be addressed individually. If it is decided that a particular stereotype would be receptive to a given approach, then the marketing campaign can target all the customers associated with the stereotype. This approach allows for cost effective targeting of the right subset of customers whilst allowing the company management to deal with a manageable number of stereotypes \cite{sarstedt2011concise}.

The growth in data provides interested parties, such as utility companies, with the ability to direct their behavioural change interventions in a more targeted way then previously. Clustering techniques provide a means of finding a few useful groupings of households where each grouping can then be targeted in an effective way particular to the characteristics of that grouping.

One behavioural trait that is useful to analyse is the flexibility that households exhibit in their daily behaviour. For instance, some households will be creatures of habit and will eat their evening meal at almost the same time each evening, whilst others have a much more variable lifestyle and will eat at varying times. Clustering households using their degree of flexibility provides a way of identifying the subset of electricity users who may be most receptive to an intervention intending to alter the timings of their activities.

A review of clustering of electricity load profiles has been published by Chicco \cite{chicco2012overview} and this shows that consideration of the flexibility features of households has not been considered previously.

\section{Methodology}
The ongoing North East Scotland energy monitoring project (NESEMP) is examining the relationship between different types of energy feedback and psycho-social measures including individual environmental attitudes, household characteristics, and everyday behaviours.  As part of this project, several hundred households were monitored and the electricity usage was recorded every five minutes using CurrentCost monitors over a period of a year \cite{Craig2012}. The volume of data input to the clustering exercise described in this abstract is of the order of 17 million individual readings.

After cleaning of households with insufficient readings, the data for 180 randomly selected households are loaded into a MySQL database and the readings are aligned with exact 5 minute boundaries (e.g. 1pm, 1.05pm, etc.) by interpolation between the actual readings. Each day of sampling is classified in a number of ways such as "working day" or "summer".

A subset of the data for the peak period of 4pm to 8pm and for working days throughout the year is extracted for further analysis. Average usage for each household within the evening period is calculated allowing for the ranking of all the households by the amount of electricity used per evening period.

\begin{figure}[h!]
\centering
\includegraphics[width=2.5in]{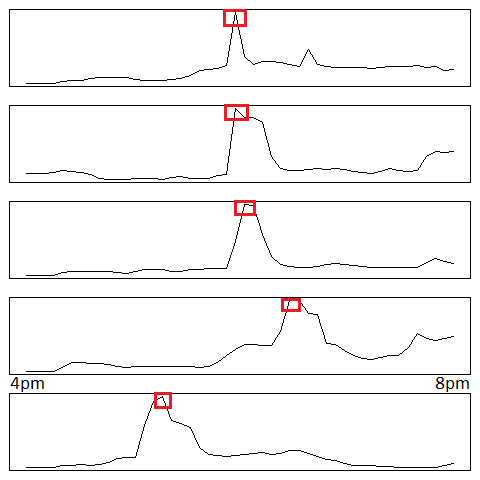}
\caption{Maximum usage at similar times each day}
\label{fig:same}
\end{figure}

\begin{figure}[h!]
\centering
\includegraphics[width=2.5in]{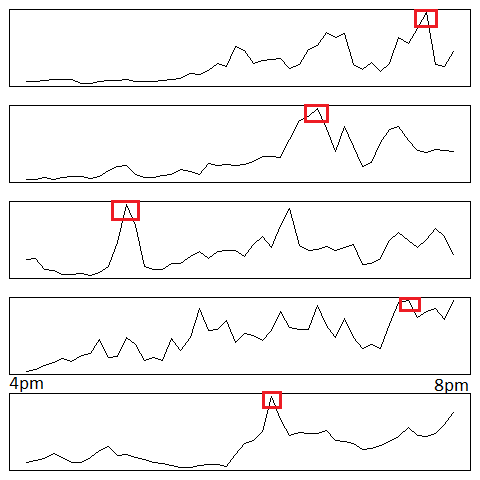}
\caption{Maximum usage at irregular times}
\label{fig:different}
\end{figure}

One measure of flexibility is, for each day and for each household in the sample, to find the time at which the maximum usage of electricity occurred during the peak period of 4pm to 8pm. This time (calculated as minutes after 4pm) for each household can then be measured and the standard deviation of this time derived as the flexibility measure. A household showing regular behaviour for an example week is given at Figure \ref{fig:same}. This suggests the members of the household arrive home at a similar time on each weekday and immediately cook their evening meal. This happens about 6pm each day with Thursday being slightly later and Friday being earlier. In comparison, Figure \ref{fig:different} shows another household for the same week and shows irregular timings for the household's evening activities.

The attributes of peak time usage and variability in time of maximum usage are used as input to a kmeans clustering exercise in order to identify four clusters. This clustering is run a number of times in order to allow for random starting conditions with the best overall solution taken.

\section{Results}

\begin{figure}[h!]
\centering
\includegraphics[width=3.3in]{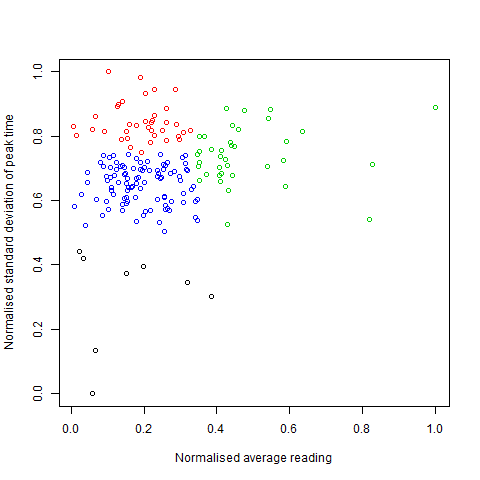}
\caption{4 clusters found using 2 attributes}
\label{fig:2attrs}
\end{figure}

The clusters derived from the kmeans algorithm are shown at Figure \ref{fig:2attrs}.

Interpretation of this simple graph shows that the households represented by green are those with higher evening usage of electricity and should be the ones targeted for general initiatives intended to reduce total electricity usage. 

Those households represented in red are the ones with the most variability in their day to day behaviour and may be the best to target with initiatives intended to change behaviour.

The households represented in black show little variability in behaviour and relatively low usage of electricity and it is likely they should be omitted from any initiatives.

Those households represented as blue are those with lower total usage and a middling degree of variability and would probably be addressed as a lower priority after the red and green households.

\begin{figure}[h!]
\centering
\includegraphics[width=3.3in]{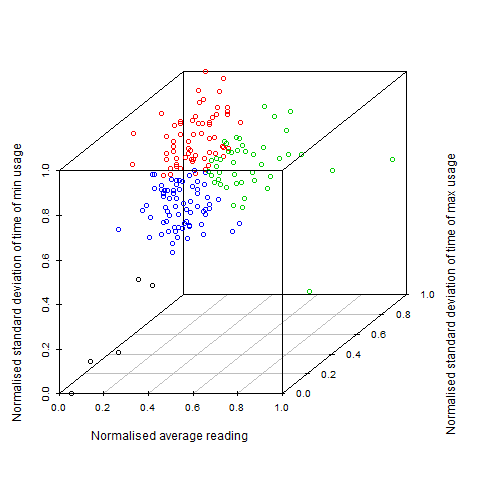}
\caption{4 clusters found using 3 attributes}
\label{fig:3attrs}
\end{figure}

Using just 2 dimensions it is simple to plot the households by their dimensions and to draw a cut off line (e.g. 50 households to the top right of the diagram) in order to select a sub group for targeting. However, adding further attributes to represent aspects of the households' behaviour will greatly complicate the selection of a subset and the application of the clustering technique becomes more important. Adding an additional attribute representing the variability in the time of minimum usage during the evening peak period leads to the clusters shown on Figure \ref{fig:3attrs}. As can be seen from the Figure, selecting a subset visually becomes more challenging and adding further attributes will increase the complexity and, thus, the clustering approach becomes more useful.

\section{Conclusions}

This initial work shows that useful clusters using the concept of flexibility can be derived and can be used in cost effectively targeting appropriate behavioural change interventions at those households most likely to respond. For example, the households showing high variability in their behaviour could be targeted with a variable pricing initiative to encourage usage of cheaper electricity at off-peak times. These households are unlikely to see this as too much disruption from their normal, very variable, behaviour pattern and are likely to react positively to the initiative. In contrast, those households showing stable habitual behaviour would probably find any overt initiative to be too disruptive to their lifestyle and can be addressed using technological solutions, such as in-house batteries, which would allow shifting of their demand on the electricity network without disruption to their daily behaviour.

\begin{figure}[h!]
\centering
\includegraphics[width=2.5in]{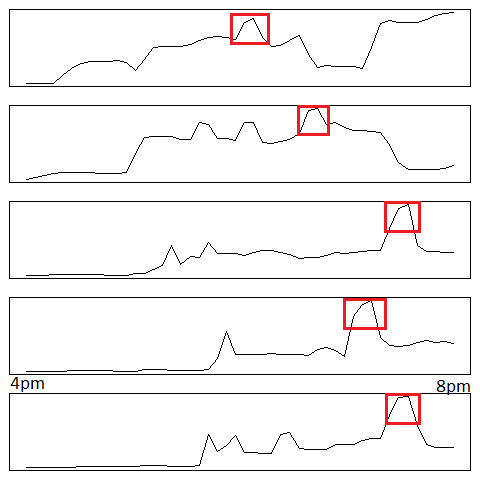}
\caption{Possible motifs}
\label{fig:motifs}
\end{figure}

Future work will expand on the concept of flexibility to explore the identification of repeating activities (e.g. retiring to bed) by making use of the pattern of electricity usage and then exploring how those activities vary daily. Motifs representing regular activities, such as cooking, will be identified within a household's meter data and the variation of the timing of the motif between days will provide a basis for clustering the households. Figure \ref{fig:motifs} shows an example week of peak time usage for a household and highlights the possible motifs.

The groupings of households created using only the meter data will be compared with the groupings determined within the NESEMP project using the demographic and attitudinal data in order to provide validation of the effectiveness of the clustering technique described.

There is an increasing interest in the effectiveness of different forms of energy feedback on household energy behaviours. It would be interesting to explore the extent to which energy-flexibility could be fed back to households in an easy-to-understand manner.

\section{Acknowledgements}

This work was possible thanks to EPSRC grant references EP/I000496/1 and EP/G065802/1.

%
\bibliographystyle{abbrv}
\bibliography{References}

\begin{thebibliography}{1}

\bibitem{chicco2012overview}
G.~Chicco.
\newblock {O}verview and performance assessment of the clustering methods for
  electrical load pattern grouping.
\newblock {\em Energy}, Volume 42, Issue 1:68–80, June 2012.

\bibitem{Craig2012}
T.~Craig, C.~Galan-Diaz, S.~Heslop, and J.~Polhill.
\newblock {T}he {N}orth {E}ast {S}cotland {E}nergy {M}onitoring {P}roject
  ({NESEMP}).
\newblock In {\em {W}orkshop on {C}limate {C}hange and {C}arbon {M}anagement}.
  The James Hutton Institute, March 2012.

\bibitem{Energy2009}
DECC.
\newblock {Towards a Smarter Future, Government Response to the Consultation on
  Electricity and Gas Smart Metering}.
\newblock 2009.

\bibitem{Kiprakis2011}
A.~Kiprakis, I.~Dent, S.~Djokic, and S.~McLaughlin.
\newblock {M}ulti-scale {D}ynamic {M}odeling to {M}aximize {D}emand {S}ide
  {M}anagement.
\newblock In {\em IEEE Power and Energy Society Innovative Smart Grid
  Technologies Europe 2011, Manchester, UK}, 2011.

\bibitem{river2005primer}
River.
\newblock {P}rimer on demand-side management with an emphasis on
  price-responsive programs.
\newblock {\em prepared for The World Bank by Charles River Associates, Tech.
  Rep}, 2005.

\bibitem{sarstedt2011concise}
M.~Sarstedt and E.~Mooi.
\newblock {\em {A} concise guide to market research: {T}he process, data, and
  methods using {IBM} {SPSS} statistics}.
\newblock Springer Verlag, 2011.

\end{thebibliography}
%
%
\balancecolumns

\end{document}